\documentclass[fleqn,10pt]{wlscirep}

\usepackage[english]{babel} 
\usepackage[utf8]{inputenc} 
\usepackage[T1]{fontenc} 

\usepackage{algorithm}
\usepackage[noend]{algpseudocode}
\usepackage{wrapfig}
\usepackage{pifont}
\usepackage{todonotes}
\usepackage{xcolor}

\title{CzechLynx: A Dataset for Individual Identification and Pose Estimation of the Eurasian Lynx}

\author[1,2]{Lukas Picek} 

\author[6,7]{Elisa Belotti} 
\author[3,4]{Michal Bojda} 
\author[7]{Lud\v{e}k Bufka} 
\author[5]{Vojtech \v{C}ermak} 
\author[3,4]{Martin Dula} 
\author[4]{Rostislav Dvo\v{r}\'{a}k}  
\author[4]{Luboslav Hrd\'{y}}  
\author[1]{Miroslav Jirik}  
\author[4]{V\'{a}clav Kocourek} 
\author[4,7]{Josefa Krausov\'{a}} 
\author[3,4]{Ji\v{r}\'{i} Labuda} 
\author[1]{Jakub Straka} 
\author[4]{Lud\v{e}k Toman} 
\author[4]{Vlado Trul\'{i}k} 
\author[4]{Martin V\'{a}\v{n}a} 
\author[3,4]{Miroslav Kutal} 

\affil[1]{Faculty of Applied Sciences, University of West Bohemia in Pilsen, Czechia}
\affil[2]{Inria, LIRMM, University of Montpellier, France}
\affil[3]{Department of Forest Ecology, Faculty of Forestry and Wood Technology, Mendel University in Brno, Czechia}
\affil[4]{Friends of the Earth Czech Republic, Carnivore Conservation Programme, Czechia}
\affil[5]{Center for Machine Perception, Czech Technical University in Prague, Czechia}
\affil[6]{Faculty of Forestry and Wood Sciences, Czech University of Life Sciences Prague, Czechia}
\affil[7]{Department of Research and Nature Protection, \v{S}umava National Park
Administration, Czechia}

\begin{abstract}
We introduce \href{https://www.kaggle.com/datasets/picekl/czechlynx}{CzechLynx}, the first large-scale, open-access dataset for individual identification, pose estimation, and instance segmentation of the Eurasian lynx (\textit{Lynx lynx}). CzechLynx contains {39,760} camera trap images annotated with segmentation masks, identity labels, and 20-point skeletons and covers {319} unique individuals across 15 years of systematic monitoring in two geographically distinct regions: southwest Bohemia and the Western Carpathians. 
{In addition to the real camera trap data, we provide a large complementary set of photorealistic synthetic images and a Unity-based generation pipeline with diffusion-based text-to-texture modeling, capable of producing arbitrarily large amounts of synthetic data spanning diverse environments, poses, and coat-pattern variations.
To enable systematic testing across realistic ecological scenarios, we define three complementary evaluation protocols: (i) geo-aware, (ii) time-aware open-set, and (iii) time-aware closed-set, covering cross-regional and long-term monitoring settings.}
{With the provided resources, CzechLynx offers a unique, flexible benchmark for robust evaluation of computer vision and machine learning models across realistic ecological scenarios.}
\vspace{-5mm}
\end{abstract}
\begin{document}

\flushbottom
\maketitle
%
%
\thispagestyle{empty}

\section{Background \& Summary}

Understanding species movement, population structure, and habitat connectivity is critical for effective conservation, especially for wide-ranging, protected carnivores \cite{rabinowitz2010range,kubala_factors_2024}. In Europe, the Eurasian lynx (\textit{Lynx lynx}) exemplifies these challenges, being ecologically important, legally protected, and indicative of broader biodiversity concerns. 
This medium-sized, solitary, and territorial carnivore lives at low population densities, with substantial home ranges estimated at 443.36$\pm$283.14 km$^2$ for males and 191.92$\pm$116.34 km$^2$ for females in Europe \cite{kubala_factors_2024,ripari2022human}.
Although primarily inhabiting forested areas and rugged terrains, the lynx demonstrates notable adaptability by using refuge areas to mitigate human pressures \cite{Oeser2023} and can disperse across Central Europe's cultural landscapes, covering distances greater than 98 km from source populations \cite{Gajdarova2021}. 
{Monitoring such wide-ranging, low-density populations across heterogeneous landscapes requires extensive, long-term effort, and camera trap networks have become a key tool for this purpose.} As a result, long-term camera trap monitoring has generated rich data that are valuable not only for ecological research, such as estimating population densities and demographic parameters \cite{Weingarth2012,Pesenti2013,Dula2021,Palmero2021}, but also for the development and evaluation of novel machine learning methods \cite{norouzzadeh2018automatically,schneider2019past}. 
{This tight connection creates a mutually beneficial loop in which ecological monitoring provides the data that drives advances in machine learning, and these advances, in turn, enable more accurate and efficient processing of ever larger and more complex conservation data. Consolidating and standardizing these datasets for machine learning is, therefore, key to unlocking their full potential for both domains and have already been taken up in several recent studies built on the data described here.} \\

\noindent{\textbf{Straka et al.} \cite{straka2024hitchhiker} used a subset of the CzechLynx dataset as a case study for 2D pose estimation of endangered species. They treated Eurasian lynx camera trap images as a testbed to compare pre-trained backbones, fine-tuning strategies, data augmentation schemes, and different mixes of real and synthetic training data, using HRNet-w32 \cite{WangSCJDZLMTWLX19} and related architectures as base models. Their experiments reveiled how real and synthetic data can be combined to support pose estimation under data-limited conditions. \\}

\noindent\textbf{Picek et al.} \cite{picek2024animal} used the part of the CzechLynx dataset from Friends of the Earth Czech Republic (FoE CZ) to evaluate a method for individual animal identification, which models foreground and background information separately. They focused on the challenges of identifying lynx individuals over time and across different locations, using data from the 15 years of camera trap monitoring in Central Europe. The dataset enabled them to demonstrate how incorporating spatial and temporal priors, along with their Per-Instance Temperature Scaling (PITS), improves identification accuracy, particularly when animals appear in new areas or exhibit changes in appearance over time. \\

\noindent\textbf{Dula et al.} \cite{Dula2021} and \noindent\textbf{Palmero et al.} \cite{Palmero2021} used some parts of the CzechLynx dataset to estimate population densities by spatial capture-recapture models and other demographic parameters of the Eurasian lynx in the West Carpathians in the Czech-Slovakia borderland and in the Bohemian Forest Ecosystem: Bavarian Forest National Park in Germany and the \v{S}umava National Park in the Czech Republic. By identifying lynx individuals, Dula et al. \cite{Dula2021} estimated apparent survival, transition rates, and population turnover. Fluctuating densities and high turnover rates indicated human-caused mortality, which could limit population growth and the dispersal of lynx to other adjacent areas, thus undermining the favourable conservation status of the Carpathian population. \\

\noindent{\noindent\textbf{Adam et al.} \cite{adam2025overview,picek2025overview} used the CzechLynx dataset as part of the \href{https://www.imageclef.org/AnimalCLEF2025}{\textit{AnimalCLEF 2025}} competition on individual animal identification. Their goal was to establish a standardized benchmark for evaluating algorithms under realistic open-set conditions across multiple species. The CzechLynx dataset was chosen because its scale, temporal depth, and natural visual diversity provide a realistic and challenging testbed for evaluating the generalization and robustness of identification models. Its use in the challenge demonstrated the dataset’s suitability for large-scale benchmarking and its potential to support community-driven development of transferable wildlife monitoring methods.

\subsection{Related Work}

{Animal-focused image datasets are now well cataloged and serve as a central resource for developing and evaluating methods for individual identification of animals across species \cite{vcermak2024wildlifedatasets}.
Available resources cover a wide range of taxa, including large terrestrial mammals (e.g., zebras, elephants), marine species (e.g., belugas, whale sharks, turtles), and carnivores (e.g., leopards, hyenas, tigers).
However, geographical coverage is not evenly distributed: most publicly available datasets originate from North America and Africa, while European resources remain relatively scarce, in part because of data privacy and conservation constraints. Large carnivores on EU red lists are particularly underrepresented in open, image-based identification datasets.}
A summary of non-public datasets is provided in Table \ref{table:private_datasets}, and public datasets with their key statistics are listed in Table \ref{table:public_datasets}. Short descriptions of selected public datasets are provided below.

\begin{wraptable}{r}{0.275\textwidth} 
\vspace{-4mm}
    \centering
    \begin{tabular}{@{}lrr@{}} %
    \toprule
     \multicolumn{2}{c}{\textbf{Individuals}}& \textbf{Images} \\
    \midrule
    Bears\cite{clapham2020automated}         & 132 & 4,674  \\
    Dogs\cite{moreira2017my}                 & 60 & 625 \\ 
    Jaguars 1\cite{timm2018large}            & 16 &  176 \\ 
    Lynxes\cite{thornton2015spatially}       & 51 & 252  \\ 
    Polar bears 2\cite{prop2020identifying}  & 15 & 42  \\ 
    Ocelots\cite{nipko2020identifying}       & \textit{N/A} & 503  \\
    Jaguars 2\cite{nipko2020identifying}     & \textit{N/A} & 680  \\ 
    Cheetahs\cite{kelly2001computer}         & \textit{N/A} & \textit{N/A}   \\ 
    Polar bears 1\cite{anderson2010computer} & \textit{N/A} & \textit{N/A}  \\ 
    \bottomrule
    \end{tabular}
    \caption{
    Non-public animal re-id datasets with large carnivores.
    }
    \label{table:private_datasets}
    \vspace{-2mm}
\end{wraptable}

The \textbf{ATRW} dataset \cite{li2019atrw} {contains} 8,076 videos collected from ten zoos in China and covers 92 Amur tigers (\textit{Panthera tigris}). Each image is annotated with bounding boxes, identity labels, and pose keypoints, allowing development of re-identification and pose estimation methods for controlled environments {and populations.}

The \textbf{Hyena ID 2022} and \textbf{Leopard ID 2022} datasets \cite{botswana2022}, developed by the Botswana Predator Conservation Trust, contain 3,129 images of 256 individual spotted hyenas (\textit{Crocuta crocuta}) and 6,795 images of 430 African leopards (\textit{Panthera pardus}), respectively.
Images were taken in natural settings and annotated with bounding boxes and viewpoints,  making these datasets valuable for testing in-the-wild generalization performance for {identification of} individual animals.

The \textbf{LionData} dataset \cite{dlamini2020automated} was compiled as part of the Mara Masai project in Kenya and includes 750 images of 98 lions (\textit{Panthera leo}). Each image was manually annotated to support training and evaluation of animal re-identification models under natural variability, such as lighting changes and different camera angles.

The \textbf{PolarBearVidID} dataset \cite{zuerl2023polarbearvidid} covers 13 individual polar bears recorded in 6 German zoos. It contains approximately 138k images extracted from 1,431 video sequences. To reduce the risk of background overfitting caused by fixed camera positions, the individuals were cropped from the original photographs.

\begin{table*}[!h]
\setlength{\tabcolsep}{0.5em} 
\centering
\begin{tabular}{llrrccccc}
\toprule
                                            \textbf{Name} & \textbf{Species} & \textbf{Records} & \textbf{Individuals} & \textbf{Span} & \textbf{BBoxes} & \textbf{Masks} & \textbf{Pose} & \textbf{Wild} \\
\midrule
PolarBearVidID \cite{zuerl2023polarbearvidid} & \textit{Ursus maritimus}& 1,431$^\dagger$ &~\,13 & unknown & $\bullet$ & $\circ$ & $\circ$  & $\circ$ \\
LionData \cite{dlamini2020automated}          & \textit{Panthera leo}    &   750~\,       &~\,94 & unknown & $\circ$ & $\circ$ & $\circ$  & $\bullet$ \\
Hyena ID 2022 \cite{botswana2022}             & \textit{Crocuta crocuta} & 3,129~\,       &  256 & unknown & $\bullet$ & $\circ$ & $\circ$  & $\bullet$ \\
Leopard ID 2022 \cite{botswana2022}           & \textit{Panthera pardus} & 6,806~\,       &  430 & unknown & $\bullet$ & $\circ$ & $\circ$  & $\bullet$ \\
ATRW \cite{li2019atrw}                        & \textit{Pantheria tigris}& 8,076$^\dagger$&~\,92 & 1 day & $\bullet$ & $\bullet$ & $\bullet$  & $\circ$ \\

\midrule
                                 (our) CzechLynx$^\ddagger$ \cite{picek_czechlynx} & \textit{Lynx lynx}    &  {39,760}~\,         & {319}
                                  & 15 years & $\bullet$ & $\bullet$ & $\bullet$  & $\bullet$ \\
\bottomrule
\end{tabular}
\caption{\textbf{All publicly available datasets with large carnivores.}
{For each dataset, we report the number of images, unique individual identities (Ids), temporal span (Span), 
and the availability of bounding boxes (BBoxes), segmentation masks (Masks), pose keypoints (Pose), 
and in-the-wild imagery (Wild). A filled bullet ({\small$\bullet$}) indicates that the annotation type is provided, 
while an open circle ({\small$\circ$}) indicates it is not.} 
$^\dagger$ Datasets derived from video recordings. $^\ddagger$ Statistics for CzechLynx exclude synthetic images.}
\label{table:public_datasets}
\end{table*}

\section{Methods}

This section {describes how the CzechLynx dataset was built}, including detailed descriptions of data acquisition, annotation, and pre-processing steps. The dataset combines data collected from long-term camera trapping surveys across two distinct regions with synthetic samples generated to enhance variability. Where applicable, previously published methods are cited, and only novel technical processes are described in detail.

\subsection{Dataset Collection}

The CzechLynx dataset \cite{picek_czechlynx} comprises data collected from two independent conservation projects operating within distinct regions of the Eurasian lynx’s (\textit{Lynx lynx}) Central European range. Each project used its own field methodology and data management pipeline. To maintain transparency and reproducibility, all data sources {are described} separately below.

\medskip
\noindent\textbf{Carnivore Conservation Programme, Friends of the Earth Czech Republic (FoE CZ), and Department of Forest Ecology, Mendel University in Brno}: 
This source includes data collected through long-term monitoring in the Western Carpathians (since 2009) and Southwest Bohemia (since 2015){; labeled as} ``\textit{FoE CZ -- The Western Carpathians}'' and ``\textit{FoE CZ -- Southwest Bohemia}'' in the following text.
Camera traps were deployed on forest roads, trails, mountain ridges, as well as marking and resting sites, identified via snow-tracking, signs of occurrence, or telemetry research \cite{dula2023first, kubala_factors_2024}. 
From 2015 onward, the Western Carpathians area followed a systematic camera trapping and spatial capture-recapture model to estimate lynx densities and evaluate density fluctuations, apparent survival, transition rate, and individual turnover during five consecutive seasons \cite{Dula2021}. Earlier data and those collected by opportunistic camera trapping allowed for estimation of minimum population size, social structure, and evidence of reproduction. Cameras used included both white-flash and infrared models (video-enabled). Most individuals were manually identified based on distinctive coat patterns visible on their flanks, forelimbs, and hindlimbs \cite{Dula2021}.

\medskip
\noindent\textbf{\v{S}umava National Park Administration}: 
This source includes data collected from the areas of \v{S}umava National Park and the Protected Landscape Area in southwestern Bohemia. The data were gathered by the \v{S}umava National Park Administration within two long-term international monitoring projects. The first project, started in 2009, covers roughly the northwestern two-thirds of \v{S}umava National Park. Its main goal is to obtain yearly density estimates of independent lynx individuals living in the core area of the Bohemian-Bavarian-Austrian lynx population.
Within the scope of the first project, data have been collected using a 2.7$\times$2.7 km grid. White-flash camera traps were installed seasonally in every second grid cell at suitable locations along forest paths, roads, and trails. Additional details about this methodology can be found in studies by Weingarth et al. \cite{weingarth2012first} and Palmero et al. \cite{palmero2021demography}.
The second project (initiated in 2017) aims to provide as complete an overview as possible of the entire Bohemian-Bavarian-Austrian lynx population throughout its full range. Data for this project were collected year-round at various sites, including forest paths, roads, trails, and lynx scent-marking locations, using both white-flash and infrared video-enabled camera traps. These cameras were set up continuously based on a 10$\times$10 km grid (ETRS LAEA 5210). In each grid cell predominantly within the \v{S}umava National Park and Protected Landscape Area, between 4 to 8  camera sites were continuously monitored throughout the year. Further information about this project is available in a separate report \cite{wolfl2020lynx}.

\subsection{Raw Data Pre-processing}

The pre-processing of raw data started with the manual removal of empty images, which frequently resulted from false triggers caused by wind, moving vegetation, or changing light conditions. Once these non-informative images were discarded, the remaining photographs were screened by a team of experts and trained volunteers, who classified the content by species, {while removing data with} humans or vehicles. Only images positively identified as containing Eurasian lynxes were used for further analysis.
{All pre-selected images then underwent a second review by specialists with long-term experience in lynx monitoring.} Using the species’ distinctive coat patterns on the hind limbs, forelimbs, and flanks, they manually identified individual lynxes. In each study area (southwest Bohemia and the Western Carpathians), at least three well-trained local {experts independently processed and cross-checked identifications in line with established camera trapping standards~\cite{Dula2021,choo2020best}.}
The resulting set of expert-verified images, each linked to a specific, individually recognized lynx, formed the basis for subsequent processing.

\subsection{Video Pre-processing}

{The CzechLynx dataset~\cite{picek_czechlynx} contains \textit{Lynx lynx} encounters collected using different camera trap models. Some devices capture still image/images, whereas others record short videos (35\% of observations). Because consecutive frames within a burst or video clip are often nearly identical, using all frames would introduce substantial redundancy and would make manual annotation time-consuming. Therefore, a small number (up to three) of the most informative frames is automatically selected for each detection event (i.e., observation).}

The selection was based on automated animal detection using MegaDetector~\cite{beery2019efficient}, a YOLOv5-based model widely used in ecological research for reliably detecting any animal in camera trap imagery\cite{fennell2022use,leorna2022human,henrich2024semi}. Each video frame was processed, and those with a detection confidence below 0.7 were discarded. From the remaining frames, the first and last with valid detections were identified, and the interval between them was divided into three equal parts. One frame {was} then selected from each segment (specifically, the frame with the largest detected bounding box) ensuring that the chosen images capture the animal prominently and are distributed across the video timeline (see Figure~\ref{fig:video_frame_selection}). 
Selected frames were used as representative samples for downstream tasks, e.g., individual identification and pose estimation. The following criteria guided the frame selection process: (i) the animal’s full body should be visible; (ii) the animal should be captured from a side view, where identifying coat patterns are most apparent; (iii) the animal should be close enough to the camera to reveal fine details without any part being cropped; and (iv) the selected frames should be temporally spaced to capture variation in pose and movement.

\begin{figure}[!t]
 \centering
 \includegraphics[width=0.975\textwidth]{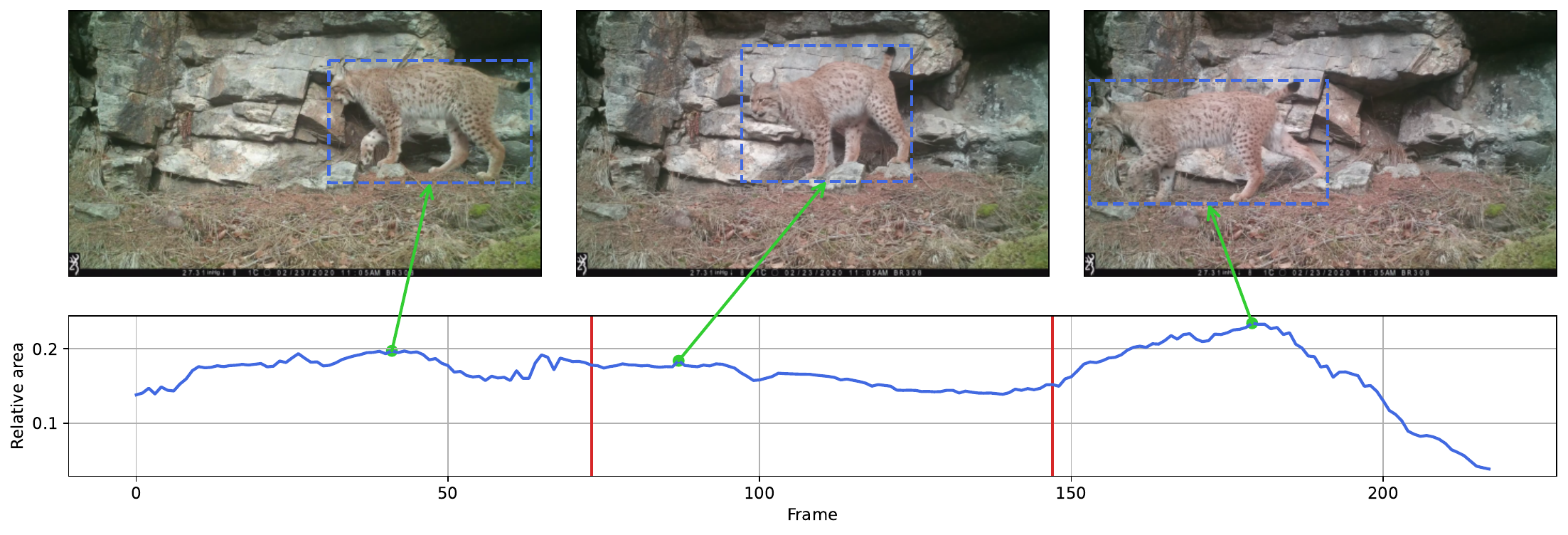}
 \caption{Frame selection based on bounding box area and temporal spacing.
 The video is divided into three parts (visualized as delimited by red lines), and one frame with the highest relative area 
 is selected from each segment.}
 \label{fig:video_frame_selection}
\end{figure}

\subsection{Dataset Annotation}

All extracted images were further processed, i.e., manually annotated, to support evaluation of three computer vision and machine learning tasks: (i) instance segmentation, (ii) individual identification, and (iii) animal pose estimation. 
The \href{https://github.com/cvat-ai/cvat}{Computer Vision Annotation Tool (CVAT)}, an open-source, web-based labeling platform, was used to create polygon, segmentation mask, keypoint, and attribute annotations.

The data for \textbf{instance segmentation} were annotated using a semi-automated, human-in-the-loop approach based on the Segment Anything Model (SAM)~\cite{kirillov2023segment}. Annotators used positive and negative points to prompt SAM to create initial masks, which were then checked and corrected (if needed) in CVAT to ensure accurate outlines of each lynx.
To better evaluate the approach, a comparison of human- and SAM-only masks with the final human-verified annotations has been done. SAM alone reached a mean Intersection over Union (IoU) of 0.96, while fully manual annotation reached an IoU of 0.87. The human-in-the-loop workflow was both faster and more accurate (30 seconds vs 5 minutes per image), as SAM helped annotators avoid small boundary errors common in manual segmentation.

The data targeted for \textbf{animal pose estimation} were also annotated using a semi-automated approach. Initial keypoint predictions were generated with a pre-trained AnimalPose model \cite{cao2019cross} and subsequently updated manually in CVAT if needed. Up to 20 key points were assigned to each individual based on the \href{https://mmpose.readthedocs.io/en/latest/demos.html}{MMPose animal skeleton} specification, covering joints, facial features, and the tail base. The detailed description of all keypoints and sample poses is provided in Figure \ref{fig:pose_samples} and Table \ref{table:keypoint_definition}. In cases of partial visibility, some key points were left unannotated if they were not visible in the image.
In addition to spatial labels, \textbf{image descriptive metadata}, including the lynx’s \textit{viewpoint} (e.g., left flank, right flank, frontal, or rear), are annotated.

\begin{figure}[!h]
   \vspace{-5mm}
 \centering
 \includegraphics[width=0.95\linewidth]{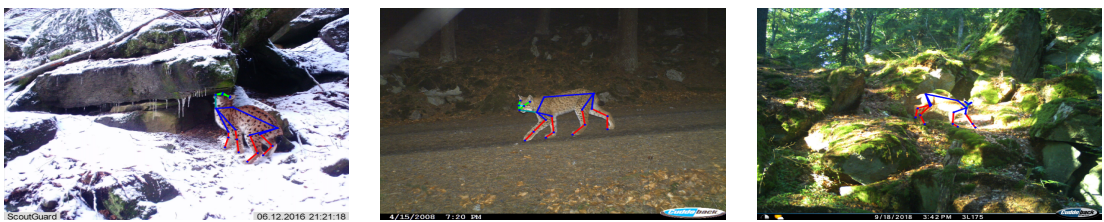}
 \caption{\textit{Lynx lynx} pose. 20-keypoint skeletal structure based on the AnimalPose~\cite{cao2019cross} standard.}
 \label{fig:pose_samples}
    \vspace{-1mm}
\end{figure}

\begin{table}[!h]
\small
\centering
\setlength{\tabcolsep}{0.55em}
\renewcommand{\arraystretch}{1.5}
\begin{tabular}{@{}|c|c|c|c|c|c|c|c|c|c|c|@{}}
\hline
 1 & 2 & 3 & 4 & 5 & 6 & 7 & 8 & 9 & 10 \\
\hline
 eye$_{\mathrm{\,left}}$ & eye$_{\mathrm{\,right}}$ & ear$_{\mathrm{\,left}}$ & ear$_{\mathrm{\,right}}$ & nose & throat & tailbase & withers & elbow$_{\mathrm{\,left}}^{\mathrm{\,front}}$ & elbow$_{\mathrm{\,right}}^{\mathrm{\,front}}$ \\
\hline
\hline
 11 & 12 & 13 & 14 & 15 & 16 & 17 & 18 & 19 & 20 \\
\hline
 elbow$_{\mathrm{\,left}}^{\mathrm{\,back}}$ & elbow$_{\mathrm{\,right}}^{\mathrm{\,back}}$ & knee$_{\mathrm{\,left}}^{\mathrm{\,front}}$ & knee$_{\mathrm{\,right}}^{\mathrm{\,front}}$ & knee$_{\mathrm{\,left}}^{\mathrm{\,back}}$ & knee$_{\mathrm{\,right}}^{\mathrm{\,back}}$ & paw$_{\mathrm{\,left}}^{\mathrm{\,front}}$ & paw$_{\mathrm{\,right}}^{\mathrm{\,front}}$ & paw$_{\mathrm{\,left}}^{\mathrm{\,back}}$ & paw$_{\mathrm{\,right}}^{\mathrm{\,back}}$ \\
\hline
\end{tabular}
\caption{{Numbered anatomical definitions of the 20 keypoints used to construct the Eurasian lynx skeleton (i.e., pose), following the AnimalPose~\cite{cao2019cross} standard and the MMPose animal skeleton definition.}}
\label{table:keypoint_definition}
\vspace{-2mm}
\end{table}

\subsection{Synthetic Data Generation} 
Synthetic data has proven valuable in improving performance in many scenarios, especially when real-world data is scarce, difficult to collect, or lacks sufficient diversity \cite{peng2015learning,straka2024hitchhiker,azizi2023synthetic}. In the context of individual animal identification or pose estimation, however, generating synthetic data that accurately captures the complexity of real-world conditions remains a significant challenge. Despite its usefulness, existing methods for generating synthetic animal pose data often struggle with realism in appearance, motion, and environmental context, which can lead to reduced accuracy in downstream applications \cite{shooter2021sydog,jiang2022prior,bonetto2023synthetic}.
To complement the real data and broaden the variability in both poses and textures, a synthetic dataset using rendered 2D images of a 3D lynx model was created. The pipeline, built with the Unity game engine, enables the creation of highly realistic synthetic samples representing multiple individuals of the \textit{Lynx lynx} species across a wide range of conditions. By leveraging detailed environmental modeling, including vegetation, terrain, and lighting, along with lifelike animations and diverse camera setups, the production of high-fidelity synthetic images that resemble real-world scenarios was ensured.

\medskip
\noindent\textbf{Data generation pipeline}: The proposed pipeline generates synthetic animal images with rich metadata, including precise pose, segmentation masks, and individual IDs. 
The process starts with building a 3D environment, populated with a rigged lynx model and realistic scenery. Four control points are manually placed in each scene, some of which are outside the camera's field of view, and the animal is animated to move between them on predefined paths. Along each path, a random set of $n$ ``stop'' points is sampled. At every stop, both the scene and the model are modified: environment variants (e.g., tree type, grass, snow coverage) and small changes to the camera viewpoint (e.g., rotation) are applied, while the model's orientation and animation state (e.g., walking, sitting) are adjusted. If multiple textures corresponding to a specific identity are available, one is randomly selected for the current frame. For each valid view, a 2D snapshot is recorded together with keypoints, bounding box, and the active texture (identity), and two samples are stored per stop (before and after modifications). Frames in which most of the model lies outside the camera's view are discarded. This procedure produces a diverse set of images that mimics real camera trap conditions. See pseudo code in Algorithm~\ref{algo:synthetic-data} for a simple overview. \vspace{-2mm}

\begin{algorithm}[H]
\small
    \caption{Synthetic Data Generation Pipeline}
    \label{algo:synthetic-data}
    \begin{algorithmic}[1]
    \small
        \Require Blank 3D mesh of animal; set of textures (each with probability $p_t$); set of animations (each with probability $p_a$)
        \Procedure{Setup}{}
            \State Setup 3D environment
            \State Prepare a set of auxiliary animations
            \State Prepare a set of textures
        \EndProcedure

        \Procedure{Simulation}{}
            \For{each 3D environment and scenario}
                \While{scenario is running}
                    \State With probability $p_t$, select a texture for the 3D mesh
                    \State With probability $p_a$, trigger auxiliary animations
                    \State Capture and store a 2D snapshot every $N$-th frame
                \EndWhile
            \EndFor
        \EndProcedure
    \end{algorithmic}
\end{algorithm}
\vspace{-7mm}

\noindent\textbf{Animal model}: A \href{https://www.turbosquid.com/3d-models/3d-model-bobcat-rigging-animation-cat/710089}{textured model} of \textit{Lynx lynx} {is used as the animal model}. The model’s skeletal structure consists of 20 keypoints and follows the Animal Pose dataset specification {, therefore,} allows for training on both real and synthetic datasets using the same model and enables direct performance comparisons.  {Since} the selected model included only a walking animation, {an additional sitting animation was manually developed to increase pose diversity.} With proper setup, other animal models can also be used.

\bigskip
\noindent\textbf{Textures}: {To increase appearance diversity between synthetic individuals, an additional set of coat textures was generated using the Paint-It diffusion-based text-to-texture model \cite{youwang2024paintit}. Such a model enables the creation of textures from short text descriptions. As illustrated in Figure~\ref{fig:textures}, the variability of the generated textures depended strongly on the choice of prompt rather than on the random seed. Therefore, a library of 28 prompts was defined by combining species terms (e.g., \textit{lynx}, \textit{bobcat}) with descriptors of common feline coat patterns (e.g., \textit{spotted}, \textit{tabby}, \textit{marbled}) and additional details. Each prompt was sampled with multiple random seeds. Despite increasing visual variability, this approach also has some practical limitations, as the diffusion model sometimes produces artifacts or unrealistic color combinations. To mitigate this, the set of prompts was iteratively refined: prompts that consistently led to low-quality textures were replaced, whereas prompts that produced stable, realistic patterns were kept. Using the final prompt set and seed combinations, 299 textures were generated and used to \textit{represent} different synthetic individuals in the simulation.}

\begin{figure}[h!]
\centering
    \includegraphics[width=\linewidth]{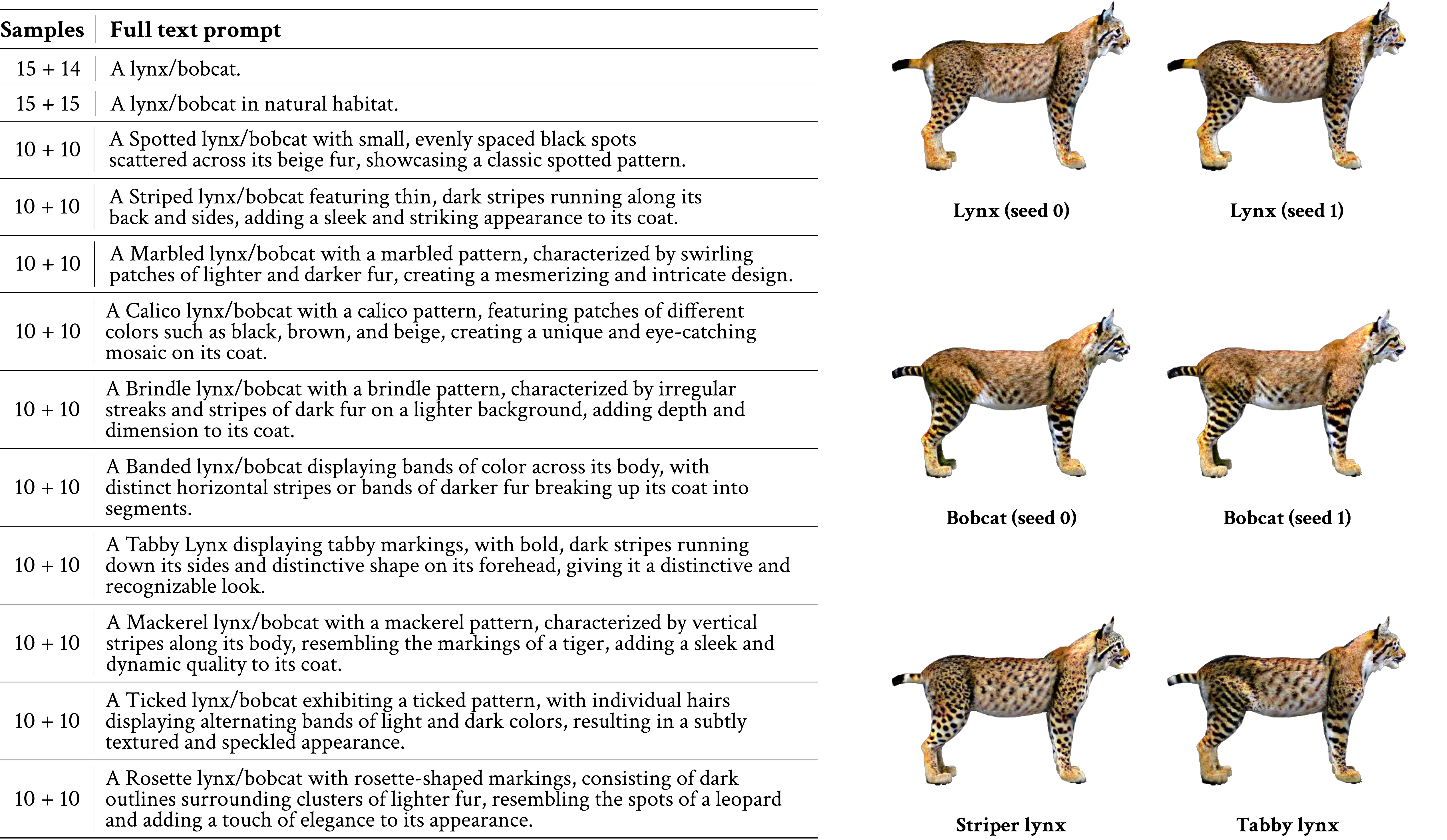}

    \caption{{\textbf{Examples of generated textures and given text prompts.} The first two rows show textures generated from the simple prompts \textit{lynx} and \textit{bobcat} with different random seeds, illustrating that changes in the seed lead to only minor visual differences for a fixed prompt, while changing the prompt already alters the overall color and pattern. The third row shows an example of a more complex prompt that results in a noticeably different texture. The first column indicates how many textures were generated for each prompt (across both \textit{lynx} and \textit{bobcat} variants and multiple seeds). In total, 299 textures were generated using 28 different text prompts.}}
    \label{fig:textures}
    \vspace{-3mm}
\end{figure}

\medskip
\noindent\textbf{Environment:} The synthetic data were designed to resemble real camera trap images closely. To achieve this, highly realistic, freely available assets from Unity’s \href{https://unity.com/demos/book-of-the-dead}{Book Of The Dead} were used, including terrain, trees, logs, and other vegetation. Several real-world scenes were selected as references, and four photorealistic scenes were recreated based on them. To enhance data variability, each scene was rendered in multiple versions that differed in tree types, grass density, and snow coverage. A comparison between real and synthetic scenes is shown in Figure~\ref{fig:synthetic_real_comparison}.

\begin{figure}[!h]
 \centering
    \includegraphics[width=0.975\textwidth]{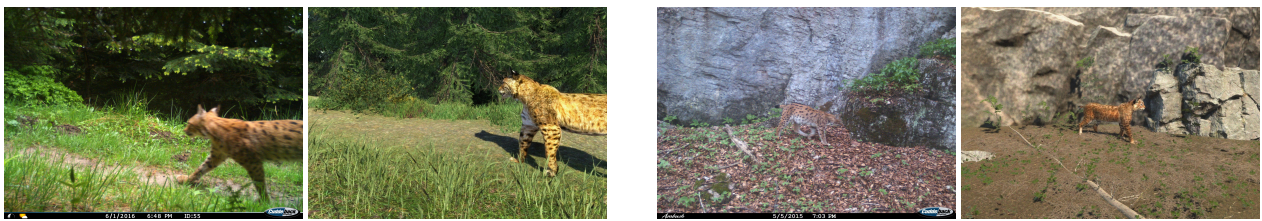}
\caption{\textbf{Selected synthetic data samples}. Inspired by the real camera trap views, we have created four highly realistic scenes. All scenes are made publicly available for further use.}
 \label{fig:synthetic_real_comparison}
\end{figure}

\section{Data Records}

The CzechLynx dataset \cite{picek_czechlynx} includes real camera trap photographs and synthetic samples of the Eurasian lynx (\textit{Lynx lynx}), organized around three downstream tasks: (i) individual identification, (ii) pose estimation, and (iii) instance segmentation.
{The main part of the dataset, consisting of 39,760 manually verified and labeled camera-trap images, is fixed, whereas the synthetic part, in practice, can be scaled to any size (for simple use, a synthetic subset with a similar number of individuals and images is provided on \href{https://doi.org/10.5281/zenodo.17592004}{Zenodo}).}
The real images span more than 15 years and come from two geographically distinct regions in Central Europe: Southwest Bohemia and the Western Carpathians. As the monitoring network expanded, the yearly volume of real images increased: fewer than 300 images were collected per year before 2012, compared to more than 5{,}000 images per year after 2020. Observations are recorded throughout the year, with the highest capture rates between January and March (40\%) and fewer images during summer months (June–August). This seasonal pattern is consistent with standard lynx monitoring designs in Central Europe, where winter camera trapping is often preferred because animals are more detectable before and during the mating season and against snow-covered backgrounds~\cite{palmero2021demography,pesenti2013density}.
Detailed statistics are provided in Table~\ref{table:dataset_sources}.

\begin{table}[!h]
\centering
\begin{tabular}{@{}lcccccc@{}}
\toprule
\textbf{Source} & \textbf{Images} & \textbf{Observations} & \textbf{Ids} & \textbf{Sites} & \textbf{{Locs}} & \textbf{Period} \\
\midrule
FoE CZ -- The Western Carpathians           & {17,997} & { 9,753} & { 95}  & { 361} & {39} & { 2009--2025}  \\
FoE CZ -- Southwest Bohemia            & \,\,{6,822}  & { 1,957}  & { 102}  & { 79}  & {32} & 2015--2023  \\
\v{S}umava National Park {Administration}            & { 14,941}  & { 7,072} & {169} & {219} & {27} & 2016--2024 \\
\midrule
Total
& {39,760} & {18,782} & {319} & {659} & {86} &{2009--2025} \\
\bottomrule
\end{tabular}
\caption{{\textbf{Summary of camera-trap sources contributing to the CzechLynx dataset.} 
For each source, we report the number of images, observation events, identified individuals (Ids), 
camera sites, spatial locations (Locs; 10$\times$10\,km grid cells), and sampling period.}
The \textit{FoE CZ -- Southwest Bohemia} and \textit{\v{S}umava National Park Administration} sources partially 
overlap geographically{, therefore, sharing 47 identities and 12 grid cells.}}
\label{table:dataset_sources}
\end{table}

All images are stored in JPEG format {(with 90\% compression)}, with metadata provided in a structured CSV file. %
{To simplify access to the data and support standardized development and evaluation of downstream tasks, i.e., instance segmentation, individual identification, and pose estimation, all real data are distributed in a single package, even though not all components are required for every task. The synthetic data are provided in the same repository in a separate archive.
Instead of maintaining separate annotation files for each downstream task, a single shared CSV file with all annotations and necessary information is provided. It contains one row per image and, for each task, indicates whether the record is used and in which split and subset (i.e., train or test).} 

\bigskip
\noindent\textbf{Metadata:} The CzechLynx dataset \cite{picek_czechlynx} includes additional metadata about the origin of the data, temporal context (e.g., observation date, relative age since first sighting, and encounter sequence), spatial context (e.g., 10$\times$10 km grid‐cell code, location and trap identifiers, and GPS coordinates), phenotypic annotation, and dataset partitioning \textit{flags} (i.e., geo‐aware, time{-aware} open, and time‐{aware} closed splits). 
These annotations support flexible filtering and grouping by identity, time, space, and experimental split, thereby enabling reproducible, domain‐aware evaluation.
The majority of metadata attributes are available for all images, observation dates are available for 98.9\% of images, location attributes for approximately 92\%, and coat-pattern annotations for about 40\%.
See Table~\ref{tab:metadata} for a detailed description of all metadata fields.

\begin{table}[!h]
\centering
\begin{tabular}{@{}p{3.5cm}p{12cm}@{}}
\toprule
\small{\textbf{Metadata}} & \small{\textbf{Description}} \\
\midrule
\textbf{\footnotesize{Source}} & \footnotesize{{Data provider}. The string \texttt{foe\_carpaths}, \texttt{foe\_bohemia}, or \texttt{snpa} correspond to FoE CZ - The Western Carpathians, FoE CZ - Southwest Bohemia, and \v{S}umava National Park, respectively.} \\
\textbf{\footnotesize{Unique name}} & \footnotesize{Unique identification of {\textit{Lynx lynx}} individual. The format is \verb+lynx_<integer>+ .}\\
\textbf{\footnotesize{Path}} & \footnotesize{Relative path to the file in the dataset.} \\
\textbf{\footnotesize{Date}} & \footnotesize{Date when the animal was observed in \texttt{yyyy-mm-dd} format. }\\
\textbf{\footnotesize{Relative age}} & \footnotesize{Relative age derived from the difference between the actual date and the first observation of the individual in the dataset.}\\
\textbf{\footnotesize{Encounter}} & \footnotesize{ID of unique sequence of images in the same camera trap location.}\\
\textbf{\footnotesize{Coat pattern}} & \footnotesize{Describes lynx's coat pattern with values \texttt{marbled} and \texttt{spotted}.}\\
\textbf{\footnotesize{Latitude, Longitude}} & \footnotesize{{WGS84 coordinates of the center of the 10$\times$10 km grid cell containing observation.}}\\
\textbf{\footnotesize{Cell code}} & \footnotesize{10$\times$10\,\,km grid‐cell identifier in the ETRS89-LAEA (EPSG:3035) pan-European coordinate system. Each entry has the form \verb+10kmE<easting_index>N<northing_index>+} \\
\textbf{\footnotesize{Location}} & \footnotesize{Unique location identifier. The closest geopolitical region to the center of the 10$\times$10\,\,km cell.} \\
\textbf{\footnotesize{Trap ID}} & \footnotesize{Unique identification of camera trap. There could be multiple in each grid cell.}\\
\midrule
\textbf{\footnotesize{Geo-aware split}} & \footnotesize{Train / test split. Images from sources with distinct populations belong to one or the other.}\\
\textbf{\footnotesize{Time-aware closed split}} & \footnotesize{Train / test split. All individuals are included in both the training and test subsets.}\\

\textbf{\footnotesize{Time-aware open split}} & \footnotesize{Train / test split. Individuals unseen in the train split are included in the test split.}\\
\textbf{\footnotesize{{Pose split}}} & \footnotesize{{Train / test split. Empty if the image is not used for pose estimation.}}\\
\midrule
\textbf{\footnotesize{{Mask}}} & \footnotesize{{Pixel-level instance segmentation mask, stored as a COCO-style RLE.}}\\
\textbf{\footnotesize{{Pose}}} & \footnotesize{{2D pose annotation, with up to 20 visible keypoints per individual stored as a dict  \{\texttt{<keypoint\_name>: [x, y]}\}; empty if no pose annotation is available.}}\\
\bottomrule
\end{tabular}
\caption{\textbf{Available metadata and their definitions}.Besides annotations, the CzechLynx dataset includes standardized information on observation identity, timing, spatial context, phenotypic annotations, and dataset splits. Spatial attributes follow a hierarchical structure: (i) WGS84 coordinates of the grid-cell centroid (\textit{coordinates}), (ii) a 10$\times$10\,\,km ETRS89-LAEA grid cell (\textit{cell code}), and (iii) the nearest administrative region to the grid-cell center (\textit{location}). For the definitions of the geo-aware, time-open, and time-closed splits, see Section \ref{splits}.}
\vspace{-3mm}
\label{tab:metadata}
\end{table}

\subsection{Task-specific subsets}
The CzechLynx dataset \cite{picek_czechlynx} is organized into three task-specific subsets designed to support the development and evaluation of (i) individual identification, (ii) pose estimation, and (iii) instance segmentation. 
{For individual identification and instance segmentation, the same set of images with clearly visible coat patterns, for which human experts confirm the identity, is provided.
Each image is paired with an identity label and a pixel-level mask outlining the lynx body, which enables the training of segmentation models while providing suitable input for re-identification.}
The pose estimation {part is} a subset of the identification/segmentation images and is smaller due to the labor-intensive annotation process. Images are selected to cover a broad range of viewpoints and behaviors and also include challenging samples with rare poses, partial occlusions, and low-visibility conditions.

\medskip
\noindent\textbf{Individual identification}: {This subset contains all real images for which individual identity can be reliably determined, together with a large synthetic complement. The real images span more than 15 years and come from two primary sources: FoE CZ (at the Western Carpathians and Southwest Bohemia) and the \v{S}umava National Park Administration (at \v{S}umava National Park and the Protected Landscape Area in southwestern Bohemia). All images are also part of the instance-segmentation subset, paired with a pixel-level mask of the lynx body. Spatial and temporal metadata are provided to enable evaluation of identity recognition models under large domain shifts, including protocols based on geo-aware and time-aware splits (see Section \ref{splits}.)

\medskip
\noindent\textbf{Instance segmentation}: {This subset uses the same real images as the one for individual identification. For all images, a binary instance mask delineating each lynx's full visible body is provided. The segmentation masks enable training and evaluation of models that separate lynx from complex natural backgrounds, including snow, forest undergrowth, and shadows. All masks are provided in COCO format using run-length encoding (RLE) and are spatially aligned with their corresponding images.} 

\medskip
\noindent\textbf{Animal pose}: {The pose-estimation subset consists of around 5k real and a large set of synthetically generated images, each annotated with up to 20 keypoints per individual (only visible keypoints are provided). The real images primarily capture side-view walking postures, reflecting the natural bias of camera trap data. Only images with a single visible individual are included, and annotations are provided in a standardized COCO format.}

\subsection{Predefined Splits}
\label{splits}

To support robust evaluation under real-world constraints, the CzechLynx dataset \cite{picek_czechlynx} is partitioned into three distinct data splits: one geo-aware and two time-aware. Each split is designed to test model generalization across spatial or temporal domains and is provided with a clear training-test split that reflects practical conservation and ecological monitoring scenarios. The summary of image counts and individual identities per split is presented in Table~\ref{table:dataset_splits}.

\medskip
\noindent\textbf{Time-aware closed-set split}:
In this split, data are divided by time and include the same individuals in both the training and test sets. Although this setup is less ecologically realistic, since wild populations are rarely closed, it provides a controlled evaluation scenario comparable to standard benchmarks used in machine learning. The data are separated using clear time cutoffs, with no samples captured within the same encounter or in temporally neighboring periods appearing in both the training and test sets, preventing train-to-test data leakage.

\medskip
\noindent\textbf{Time-aware open-set split}:
In this split, data are also divided by \textit{time}, with most individuals (82) in the test set also present in the training set. However, the test images come from later time periods, separated from the training data by season. This setup reflects realistic long-term monitoring scenarios in which populations naturally change over time (e.g., new individuals are born, some individuals die, and others may age or change appearance). The split, therefore, evaluates a model’s ability to generalize across temporal variation and population drift. The split was done using a clean time-cutoff strategy to prevent data leakage, ensuring that no images from neighboring seasons appear in both training and test sets.

\medskip
\noindent\textbf{Geo-aware split}:
This split is designed to evaluate spatial generalization. Training data come from the \textit{The Western Carpathians}, while the testing comes from geographically distinct \textit{southwest Bohemia}. The individuals in the training and test sets are completely disjoint, ensuring that no data leakage occurs. This setup reflects an ecological scenario where re-identification models will be deployed in a new area without locally labeled data, which tests generalization and transferability to different populations of the same species, but observed in other regions.

\begin{table}[!h]
\centering
\begin{tabular}{lrrrrrrrr}
\toprule
& \multicolumn{2}{c}{\textit{\# of images }} & \multicolumn{2}{c}{\textit{\# of identities}} & \multicolumn{2}{c}{\textit{\# of sites}} & \multicolumn{2}{c}{\textit{\# of {locations}}} \\
& training & test & training & test & training & test & {training} & {test} \\
\midrule
Geo-aware-open     & {21,763} & {17,997} & {224} &  {95} & {298} & {361} & {47} & {39}\\
Time-aware-open    & {27,587} &  {12,173} & {275} &  {126} & {565} & {313} & {82} & {63}\\
Time-aware-closed  & {27,836} &  {11,924} & {319} & {319} & {603} & {464} & {83} & {77}\\
\midrule
\end{tabular}
\caption{Number of identities, locations, sites, and images {across} three provided splits; two time-aware and one geo-aware.
{All splits were created to prevent training-to-test data leakage, i.e., samples that are close in time and/or location will never appear in both the training and test subsets. In the open-set split, 44 new identities appear in the test set.}}
\label{table:dataset_splits}
\end{table}
\section*{Technical Validation}

To ensure the technical quality of the CzechLynx dataset \cite{picek_czechlynx}, we performed a series of validation steps focused on the consistency, accuracy, and usefulness of the data for both computer vision and ecological research.

\medskip
\noindent\textbf{Annotation Quality Control.}
The dataset was annotated using standard tools such as CVAT. To reduce labeling errors, a dual-pass validation strategy was used. First, annotations were created or verified by trained annotators. Then, a second round of validation was performed by a separate team to ensure consistency in identity assignment, keypoint placement, and segmentation masks.

\medskip
\noindent\textbf{Data Consistency and Identity Validation.}
The subset for individual identification includes only images where individual identity could be confirmed by at least two independent observers, based on distinct coat patterns on limbs and flanks. Any ambiguous cases were excluded. This process ensured a high level of reliability in the identity labels across the dataset.

\medskip
\noindent\textbf{Synthetic Data Comparison.}
To address pose diversity limitations, a synthetic subset was generated using a 3D model and the Unity engine. We performed side-by-side comparisons of real and synthetic images (see Figure~\ref{fig:synthetic_real_comparison}) and confirmed that the synthetic data closely mirrors real-world camera trap imagery. The synthetic samples also include similar metadata and annotations, such as keypoints and segmentation masks.

\medskip
\noindent\textbf{Data Distribution and Coverage.}
The dataset spans more than 15 years and covers 319 individual lynx across two populations and multiple regions. Images were collected from 659 sites, with the majority sampled systematically in a 10$\times$10 km grid. This ensures representative coverage of habitats and individual variation. The dataset also includes time- and geographically-aware splits, enabling robust open- and closed-set evaluation.

\section*{Code Availability}

Code to load and work with the CzechLynx dataset \cite{picek_czechlynx}, including baseline training and evaluation scripts, is available in the open-source \href{https://github.com/WildlifeDatasets/wildlife-datasets}{\texttt{WildlifeDatasets}} repository. 
Executable baseline notebooks are provided on Kaggle in the \href{https://www.kaggle.com/datasets/picekl/czechlynx/code}{Code section} associated with the dataset.
The Unity-based pipeline and diffusion-based tools used to generate the synthetic images and annotations are available in a separate repository, \href{https://github.com/WildlifeDatasets/wildlife-synthetic}{\texttt{WildlifeSynthetic}}.

\section*{Data Availability}

The full dataset is publicly available on \href{https://doi.org/10.5281/zenodo.17592004}{Zenodo}. 
A mirrored copy of the images and annotations, together with example notebooks, is also hosted on  \href{https://www.kaggle.com/datasets/picekl/czechlynx}{Kaggle}.

\section*{Author contributions}

\noindent\textbf{Conceptualization and Methodology}: Lukas Picek.  

\noindent\textbf{Software}: Lukas Picek, Miroslav Jirik, Jakub Straka, and Vojtech Cermak.

\noindent\textbf{Data Acquisition}: Miroslav Kutal, Elisa Belotti, Luděk Bufka, Martin Duľa, Rostislav Dvořák, Michal Bojda,Václav Kocourek, Josefa Krausová, Jíří Labuda, Luděk Toman, Martin Váňa

\noindent\textbf{Data Curation}: Lukas Picek, Miroslav Jirik, and Jakub Straka.

\noindent\textbf{Writing – Original Draft}: Lukas Picek, Miroslav Kutal, Martin Duľa, Miroslav Jirik, Jakub Straka, Vojtech Cermak, Elisa Belotti, and Josefa Krausová.

\noindent\textbf{Writing – Review \& Editing}: All authors read and approved the final manuscript.

\section*{Competing Interests}
The author(s) declare no competing interests.
\section*{Acknowledgements}
This research was supported by the Technology Agency of the Czech Republic, project No. SS05010008. We would like to express our gratitude to Peter Drengubiak, Martina Du\v{s}ková, \v{S}árka Frýbová, Martin Gendiar, Beňadik Machciník, Michal Králik, Michal Kudlák, Leona Marčáková, Martin \v{S}pilák, Zdeněk Tyller, Gabriela Váňová and the dedicated volunteers of Carnivore Tracking Project for their help collecting data and fieldwork. 

\bibliography{bibliography}

\end{document}